\documentclass[12pt,draftclsnofoot,onecolumn]{IEEEtran}
\IEEEoverridecommandlockouts
\usepackage{cite}
\usepackage{amsmath,amssymb,amsfonts}
\usepackage{algorithmic}
\usepackage{graphicx}
\usepackage{subcaption}
\usepackage{textcomp}
\usepackage{xcolor}
\usepackage{bm}
\usepackage{amsmath}
\usepackage{multirow}
\usepackage{times}
\usepackage{caption}

\def\BibTeX{{\rm B\kern-.05em{\sc i\kern-.025em b}\kern-.08em
    T\kern-.1667em\lower.7ex\hbox{E}\kern-.125emX}}
\begin{document}

\title{
\huge Clutter Resilient Occlusion Avoidance\\for Tightly-Coupled Motion-Assisted Detection
\thanks{
This work has been accepted by 2025 IEEE International Conference on Acoustics, Speech, and Signal Processing (ICASSP 2025). This work was supported by the National Natural Science Foundation of China (Grant No. 62371444), the Shenzhen Science and Technology Program (Grant No. RCYX20231211090206005), and the Science and Technology Development Fund of Macao S.A.R (FDCT) (No. 0081/2022/A2). 

Corresponding author: Shuai Wang ({\tt\small s.wang@siat.ac.cn}).}
}
\author{
Zhixuan Xie$^{1,2}$, Jianjun Chen$^{1}$, Guoliang Li$^{3}$, Shuai Wang$^{2,\dag}$, Kejiang Ye$^{2}$, Yonina C. Eldar$^{4}$, Chengzhong Xu$^{3}$ 
\\
\vspace{0.2in}
$^{1}$ Southern University of Science and Technology \\
$^{2}$ Shenzhen Institutes of Advanced Technology, Chinese Academy of Sciences \\
$^{3}$ University of Macau \quad 
$^{4}$ Weizmann Institute of Science
}
\maketitle

\begin{abstract}
Occlusion is a key factor leading to detection failures. This paper proposes a motion-assisted detection (MAD) method that actively plans an executable path, for the robot to observe the target at a new viewpoint with potentially reduced occlusion. In contrast to existing MAD approaches that may fail in cluttered environments, the proposed framework is robust in such scenarios, therefore termed clutter resilient occlusion avoidance (CROA).
The crux to CROA is to minimize the occlusion probability under polyhedron-based collision avoidance constraints via the convex-concave procedure and duality-based bilevel optimization. The system implementation supports lidar-based MAD with intertwined execution of learning-based detection and optimization-based planning.
Experiments show that CROA outperforms various MAD schemes under a sparse convolutional neural network detector, in terms of point density, occlusion ratio, and detection error, in a multi-lane urban driving scenario.
\end{abstract}

\begin{IEEEkeywords}
Motion-assisted detection, occlusion avoidance
\end{IEEEkeywords}

\section{Introduction}

Target detection aims to collect measurements of an object of interest (i.e., target) and estimate its state (e.g., pose and shape).
Conventional detection methods \cite{wang2022federated, feng2020deep, arnold2020cooperative, yan2018second,li2021learning} focus on designing deep neural networks (DNNs) to extract features from sensor points, which may break down under low quality of sensor measurements. 
Emerging motion-assisted detection (MAD) \cite{masnavi2023vacna,masnavi2022visibility,nageli2017real,shi2023real,falanga2018pampc,wang2023active,zhang2022collision,hung2023review,han2023rda} optimizes the sensor measurements by changing the viewpoint of the detector. 
In this case, the performance of MAD depends not only on the learning-based detector, but also the optimization-based planner. 

Existing MAD can be divided into planner-oriented \cite{zhang2022collision,hung2023review,han2023rda} and detector-oriented \cite{shi2023real,falanga2018pampc,nageli2017real,masnavi2022visibility,masnavi2023vacna,wang2023active}.
Planner-oriented MAD (PMAD), e.g., path following (PF) \cite{hung2023review} and regularized dual alternating direction method of multipliers (RDA) \cite{han2023rda}, improves the detection performance by moving the robot closer to the target. 
However, this ignores the view requirement of the detector.
On the other hand, detector-oriented MAD (DMAD) \cite{shi2023real,falanga2018pampc,nageli2017real,masnavi2022visibility,masnavi2023vacna,wang2023active} exploits the view requirement via learning or optimization. Learning-based DMAD \cite{wang2023active,shi2023real,falanga2018pampc} is good at predicting future views, but suffers from a lack of interpretability, e.g., the output trajectory may violate vehicle dynamics \cite{wang2023active}.
Optimization-based DMAD, e.g., occlusion-aware model predictive control (OMPC) \cite{nageli2017real,masnavi2022visibility,masnavi2023vacna}, optimizes the viewpoints while guaranteeing executable paths. However, they ignore the obstacle geometries and adopt over-simplified occlusion, collision, and dynamics models, resulting in loosely-coupled detector-planner design and degraded performance in cluttered environments.

To fill the gap, this paper proposes clutter-resilient occlusion avoidance (CROA), which is a tightly-coupled MAD framework that exploits the inter-dependency between a learning-based detector and an optimization-based planner.  
The detector is realized via a lidar-based DNN and its performance is optimized by minimizing a weighted sum of the occlusion probability and the target-robot distance. The occlusion probability captures target uncertainties, and is represented by a set of nonconvex constraints, which are handled by a convex-concave procedure (CCP).  
The planner is realized via MPC, and it models each obstacle as a polyhedron.
This allows the robot to have a larger planning space when varying the detector viewpoint, compared to existing MAD methods that model each obstacle as a ball \cite{nageli2017real,masnavi2022visibility,shi2023real,zhang2022collision}. 
The polyhedron-based collision avoidance constraints are tackled by duality-based bilevel optimization. 

To evaluate CROA, it is necessary to adopt a high-fidelity MAD simulator with intertwined execution of learning-based detection and optimization-based planning. However, existing works \cite{nageli2017real,masnavi2022visibility,shi2023real,falanga2018pampc} support 2-dimensional (2D) MAD (e.g., Yolo \cite{nageli2017real,shi2023real}) instead of 3D MAD. We thus implement a lidar-based 3D MAD simulator based on the Car Learning to Act (Carla) platform \cite{dosovitskiy2017carla} and the robot operation system (ROS).
Results in Carla-ROS demonstrate the superiority of the proposed CROA over existing PF-assisted detection \cite{hung2023review}, RDA-assisted detection \cite{han2023rda}, and OMPC-assisted detection \cite{nageli2017real}, in a multi-lane driving scenario. Particularly, the number of target points is significantly increased, while the occlusion ratio and detection error are reduced by at least $14.6\%$ and $5\%$, respectively.

The remainder of the paper is organized as follows. Section \ref{section2} presents the problem formulation of CROA. Section \ref{section3} presents the core optimization algorithms. Simulations are demonstrated and analyzed in Section \ref{section4}. 
Finally, conclusions are drawn in Section \ref{section5}.

\section{Problem Formulation}\label{section2}

As shown in Fig.~1, we consider a robot MAD system, which consists of an ego-robot and a target, operating in an environment with $K$ obstacles. 
The robot state is viewed as discrete measurements in the time domain, where the period between consecutive states is denoted as $\Delta t$.
At the $t$-th time slot ($t \in \{1,\dots,T\}$), the robot state is $\mathbf{s}_{t}=(x_{t},y_{t},\theta_{t})$, where $\mathbf{p}_t=(x_{t},y_{t})$ and $\theta_{t}$ are associated positions and orientations, respectively. 
The bounding box of the ego robot is $\mathbb{G}_{t}(\mathbf{s}_t)$.
The point cloud of the robot lidar is $\mathbf{c}_{t}\in\mathbb{R}^{3\times C_t}$, where $C_t$ is the number of measurement points.
The robot adopts a DNN-based detector, denoted as a function $g(\cdot)$, to map $\mathbf{c}_{t}$ into a list ${\mathcal{B}}_{t}=g(\mathbf{c}_{t})$, where $\mathcal{B}_{t}= \{ \mathbb{T}_t\}\cup\{\mathbb{O}_{1,t},\cdots,\mathbb{O}_{K,t}\}$, with $\mathbb{T}_t$ and $\mathbb{O}_{k,t}$ representing the bounding boxes of target and the $k$-th obstacle, respectively.
To simplify notation, it is assumed that the $K$ obstacles are static, and $\mathbb{O}_{k,t}$ reduces to $\mathbb{O}_{k}$.

\begin{figure}[!t]
  \centering
    \centering
    \includegraphics[width=0.95\textwidth]{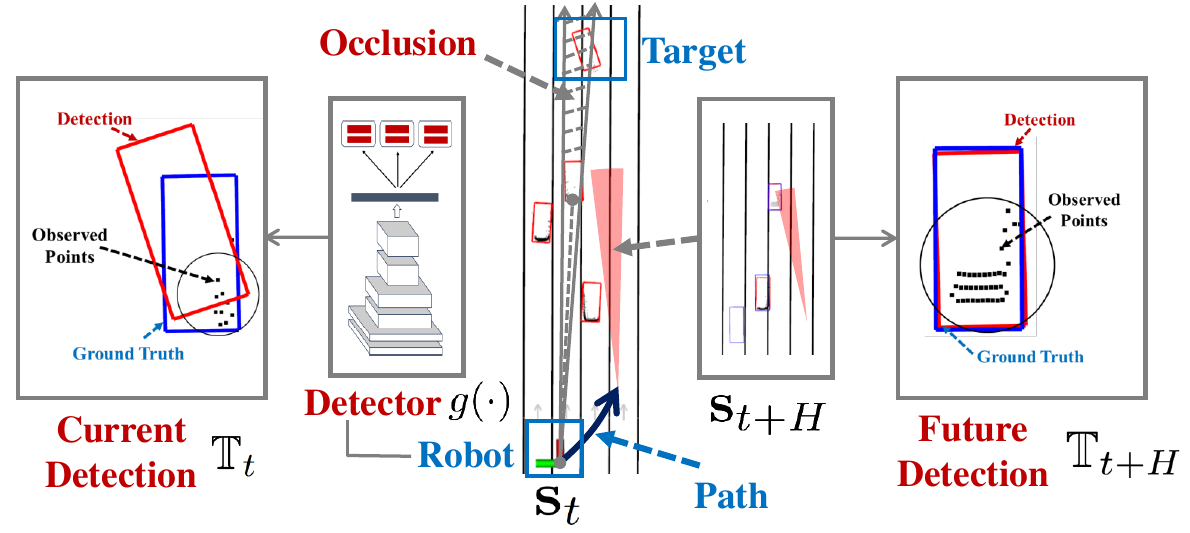}
    \captionsetup{font={small,rm}, labelfont={small,rm}}
    \caption{System model of robot MAD.}
\end{figure}

The planner adopts the MPC framework that is operated in a receding horizon fashion, where the length of prediction steps in each horizon is $H$. 
The state evolution is 
${\mathbf{s}_{h + 1}} = {\mathbf{s}_h} + f({\mathbf{s}_h},{\mathbf{u}_h})\Delta t, \ \forall h\in\mathcal{H}=\{t,\cdots,t+H\}$, where $f$ is the predefined kinematic model \cite[Sec. III-B]{han2023rda}, and $\mathbf{u}_{t}=(v_{t},\psi_{t})$ is the robot action, with $v_{t}$ and $\psi_{t}$ being the linear velocity and steering angle, respectively.
The action vector is bounded as $\mathbf{u}_{\min } \preceq \mathbf{u}_{h} \preceq \mathbf{u}_{\max },~\forall h$, where ${\mathbf{u}_{\min }}$ and ${\mathbf{u}_{\max }}$ are the minimum and maximum values of control commands. 
To simplify notations, the above planner constraints are collectively defined as $\{\mathbf{s}_h,\mathbf{u}_h\}_{h=t}^{t+H}\in\mathcal{F}$, where $\mathcal{F}$ denotes the set of all physically-feasible solutions.

The robot aims to find a path $\mathbf{s}_t\rightarrow \mathbf{s}_{t+H}$ under the planner constraints, such that the newly observed data at time $t+H$ provides a more accurate $\mathbb{T}_{t+H}$.
This is equivalent to reducing the target occlusion probability and the target-robot distance \cite{arnold2020cooperative,zhang2021distributed,nageli2017real,masnavi2022visibility} at time $t+H$.
The problem is thus
\begin{subequations}
    \begin{align}
    &\mathsf{P}:\min \limits_{\{\mathbf{s}_{h}, \mathbf{u}_{h}\}\in\mathcal{F}}C_{\text{occ}}(\mathbf{s}_{t+H})+
   \rho \, C_{\text{tar}}(\{\mathbf{s}_h\}_{h=t}^{t+H}), \label{occ_obj}  \\
&\text {s.t.}\min_{\mathbf{x},\mathbf{y}} \ \{\|\mathbf{x}-\mathbf{y}\|_2| \mathbf{x}\in\mathbb{G}_{h}(\mathbf{s}_h), \mathbf{y}\in \mathbb{O}_{k} \}
\geq d_{0},~\forall k,h, \label{collision}
    \end{align}
\end{subequations}
where $\rho$ is a weighting constant, \eqref{collision} is the collision avoidance constraint, and $d_0$ is the pre-defined safety distance. 
Cost 
$C_{\text{occ}}(\mathbf{s}_{t})$ is the target occlusion probability detailed in Section III, and $C_{\text{tar}}(\{\mathbf{s}_h\}) = 
\sum^{t+H}_{h=t} 
    \left\|\mathbf{s}_{h}-\mathbf{s}_{h}^\diamond\right\|_2^2$ is adopted to encourage the robot to approach the target, with $\{\mathbf{s}_t^\diamond,\cdots,\mathbf{s}_{t+H}^\diamond\}$ being waypoints on the naive straight line pointing towards the target direction \cite{li2024edge, han2024neupan}.

Problem $\mathsf{P}$ is nontrivial to solve due to the complex function $C_{\text{occ}}$ in \eqref{occ_obj} and the bilevel optimization in \eqref{collision}.
To solve $\mathsf{P}$, existing methods simplify $C_{\text{occ}}$ to distance-based calculation \cite{nageli2017real,masnavi2022visibility,shi2023real,tong2022environment}, and approximate $\{\mathbb{G}_{h},\mathbb{O}_{k}\}$ in \eqref{collision} as norm balls \cite{nageli2017real,masnavi2022visibility,shi2023real,zhang2022collision}, which ignore the obstacle geometries. 
In contrast, we will optimize the exact $C_{\text{occ}}$ using CCP, and the exact $\{\mathbb{G}_{h},\mathbb{O}_{k}\}$ using polyhedrons, thereby achieving potentially lower detection error. 
As for the learning-based detector, most existing methods \cite{nageli2017real,masnavi2022visibility,shi2023real,masnavi2023vacna} focus on 2D MAD (e.g., Yolo \cite{shi2023real}). Here, we consider 3D MAD that processes point clouds via DNNs and generates 3D bounding boxes.

\section{Proposed CROA Method}\label{section3}

\subsection{Occlusion Probability}\label{AA}

\begin{figure}[!t]
  \centering
    \centering
    \includegraphics[width=0.95\textwidth]{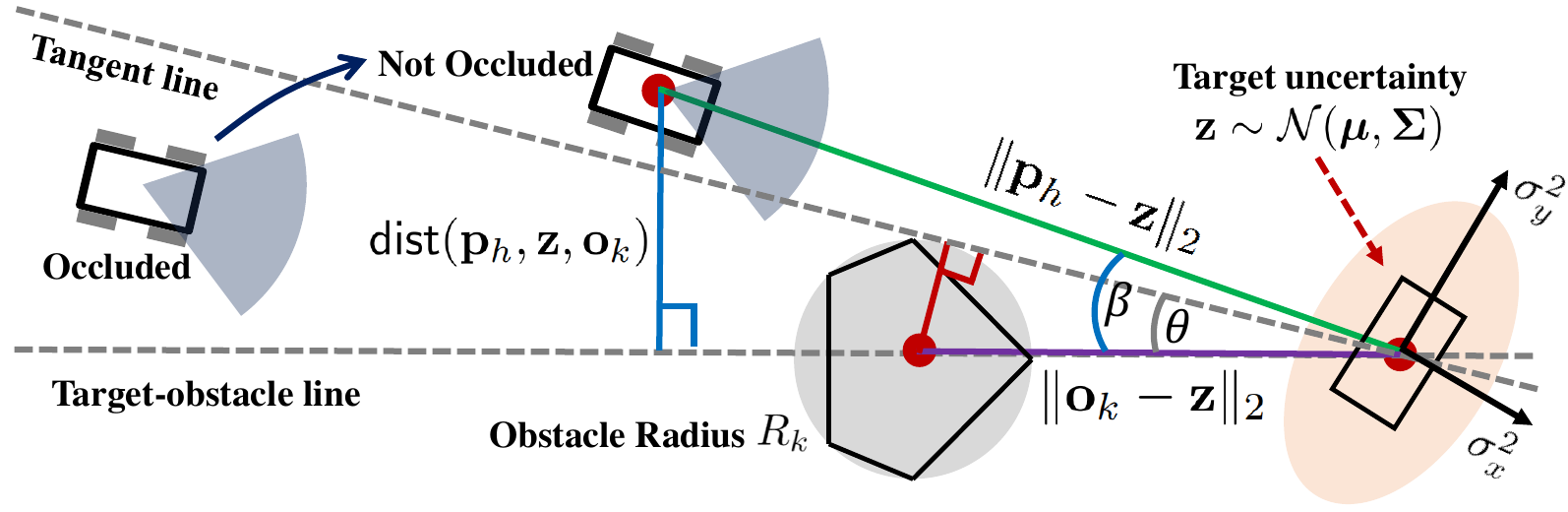}
    \captionsetup{font={small,rm}, labelfont={small,rm}}
    \caption{Occlusion probability computation of CROA.}
\end{figure}

The distribution of a target point $\mathbf{z} = (z_{x},z_{y})$ is assumed to be known and unchanged within the receding horizon, which is modelled as $\mathbf{z}\sim\mathcal{N}(\bm{\mu},\mathbf{\Sigma})$ \cite{thomas2021exact}, where $\mathcal{N}$ is the Gaussian distribution, $\bm{\mu}$ is the mean vector, and $\mathbf{\Sigma}$ is the covariance matrix. 
Its probability density function (pdf) is $\Xi(\mathbf{z})$.\footnote{In practice, the distribution can be either estimated by ego-robot using Kalman filter \cite{nageli2017real} or by other agents using cooperative perception \cite{zhang2022collision}.}
With the probability model, we adopt ray casting to compute the target visibility observed from the robot at the $h$-th time step. 
Specifically, the robot is occluded by the $k$-th obstacle if the opening angles $\beta$ and $\theta$ in Fig.~2 satisfy $\sin\theta \geq \sin\beta$.\footnote{It is assumed that the $k$-th obstacle is between the robot and the target at the $t$-th frame \cite{tong2022environment,nageli2017real}.}
According to geometric calculation, we have 
$\sin\theta=R_{k}/\|\mathbf{o}_{k}-\mathbf{z}\|_2$, where $R_{k}$ is radius of the $k$-th obstacle.
Furthermore, $\sin\beta=\mathsf{dist}(\mathbf{p}_h,\mathbf{z},\mathbf{o}_{k})/\|\mathbf{p}_h-\mathbf{z}\|_2$, where $\mathbf{o}_{k} = (o_{x,k},o_{y,k})$ is the center position of $\mathbb{O}_{k}$, and 
\begin{align}
    \mathsf{dist}(\mathbf{p}_h,\mathbf{z},\mathbf{o}_{k}) &= 
\frac{\left|-\frac{z_{y}-o_{y,k}}{z_{x}-o_{x,k}} (x_{h} - o_{x,k}) + y_{h} - 
    o_{y,k}\right|}{\sqrt{\left(\frac{z_{y}-o_{y,k}}{z_{x}-o_{x,k}}\right)^2 + 1}},
    \label{dist}
\end{align}
which is the minimum distance between the ego robot and the line-of-sight path between target and obstacle.
Therefore, $\sin\theta \geq \sin\beta$ reduces to
\begin{align}
    \frac{R_{k}}{\|\mathbf{o}_{k}-\mathbf{z}\|_2}
    \geq
    \frac{\mathsf{dist}(\mathbf{p}_h,\mathbf{z},\mathbf{o}_{k})}{\|\mathbf{p}_h-\mathbf{z}\|_2}
. \label{occ2}
\end{align}
Let $\mathcal{Z}$ denote the set of $\mathbf{z}$ satisfying \eqref{occ2} for all $k$, which is a function of $\mathbf{p}_h$, thus $\mathbf{s}_h$.
The expected occlusion probability at time $t+H$ is thus
\begin{align}
&C_{\text{occ}}(\mathbf{s}_{t+H}) = 
\int
\mathbb{I}_{\mathcal{Z}(\mathbf{s}_{t+H})}(\mathbf{z})
\Xi(\mathbf{z})
\, d\mathbf{z},
\end{align}
where indicator function $\mathbb{I}_{\mathcal{Z}}(\mathbf{z})=1$ if $\mathbf{z}\in\mathcal{Z}$ and $\mathbb{I}_{\mathcal{Z}}(\mathbf{z})=0$ otherwise.

\subsection{CCP-Based Method}

To handle the integration, we adopt Gaussian randomization \cite{luo2010semidefinite} that generates $M$ random samples $\{\mathbf{g}_1,\cdots,\mathbf{g}_M\}$ with $\mathbf{g}_i\sim\mathcal{N}(\bm{\mu},\mathbf{\Sigma})$.
As such, $C_{\text{occ}}$ is approximated by
\begin{align}
C_{\text{occ}}(\mathbf{s}_{t+H})\approx
\sum_{i=1}^M
Q_i\,
\mathbb{I}_{\mathcal{Z}(\mathbf{s}_{t+H})}(\mathbf{g}_i), \label{C_occ_app}
\end{align}
where 
$Q_i=\Xi(\mathbf{g}_i)[\sum_{j=1}^{M} \Xi(\mathbf{g}_j)]^{-1}$.
To further tackle the indicator function $\mathbb{I}_{\mathcal{Z}}(\cdot)$, a set of slack variables 
$\{\mathbf{w},\bm{\xi}\}$ with $\mathbf{w}=\{w_{1},\cdots,w_{M}\}$ and $\bm{\xi}=\{\xi_{1},\cdots,\xi_{M}\}$ are introduced, 
and the following reformulated problem $\mathsf{P}_1$ is obtained:
\begin{subequations} 
    \begin{align}
    \mathsf{P}_1:~&\min \limits_{\{\mathbf{s}_h, \mathbf{u}_h\}\in\mathcal{F},\{\mathbf{w},\bm{\xi}\}}~~ 
\sum_{i=1}^M
Q_i\|w_i\|_0
    +
   \rho \, C_{\text{tar}}(\{\mathbf{s}_h\}), \label{7a}\\
    \text { s.t. }~~&
      w_{i} \geq \left[\xi_{i}-1\right]^+, \ \forall i, \label{w}
  \\
    &\xi_{i}\geq
    \frac{R_k^2 \ \|\mathbf{p}_{t+H}-\mathbf{g}_i\|_2^2}{
\|\mathbf{o}_{k}-\mathbf{g}_i\|_2^2 \,
\mathsf{dist}^2(\mathbf{p}_{t+H},\mathbf{g}_i,\mathbf{o}_{k})
}, \ \forall i,k \label{xi} 
\\
&\textsf{constraint}~(\ref{collision}), 
    \end{align}
\end{subequations}
where $\|\cdot\|_0$ denotes $l_0$ norm, and the operator $[x]^+$ takes the nonnegative of $x$.
Based on the monotonic increasing property of $\|\cdot\|_0$ in $\mathbb{R}^+$, it can be proved by contradiction that the optimal $\{\mathbf{w}^*,\bm{\xi}^*\}$ to $\mathsf{P}_1$ always activates the constraints \eqref{w} and \eqref{xi}. 
By equalizing \eqref{w}--\eqref{xi} and plugging them into \eqref{7a}, we can show that $\mathsf{P}_1$ is equivalent to $\mathsf{P}$ as $M\rightarrow\infty$.

To proceed in solving $\mathsf{P}_1$, we need to tackle the nonconvexity in \eqref{xi}.
First, the constraint \eqref{xi} is rewritten as 
\begin{align}
\underbrace{\frac{R_k^2\|\mathbf{p}_{t+H}-\mathbf{g}_i\|_2^2}{
    \xi_{i}
}}_{:=\Phi_{i,k}}
\underbrace{-
    \|\mathbf{o}_{k}-\mathbf{g}_i\|_2^2 \,
\mathsf{dist}^2(\mathbf{p}_{t+H},\mathbf{g}_i,\mathbf{o}_{k})
    }_{:=\Theta_{i,k}}
    \leq 0. \nonumber
\end{align}
This constraint is reformulated into a compact form:
\begin{align}
\Phi_{i,k}(\mathbf{s}_{t+H}, \xi_{i})
+
\Theta_{i,k}(\mathbf{s}_{t+H})
    \leq 0, 
    \ \forall i,k. \label{ccc}
\end{align}
The term $\Phi_{i,k}$ is convex, since it is the perspective transformation of quadratic function $R_k^2\|\mathbf{p}_{t+H}-\mathbf{g}_i\|_2^2$ over $\xi_{i}$. The term $\Theta_{i,k}$ is concave, since 
the Hessian matrix of $\mathsf{dist}$ in \eqref{dist} is positive semi-definite and thus that of 
$\Theta_{i,k}$ is negative semi-definite.
Therefore, \eqref{ccc} is a convex-concave constraint, which can be tackled by the CCP technique \cite{lipp2016variations,sun2016majorization}:
\begin{align}
\Phi_{i,k}(\mathbf{s}_{t+H}, \xi_{i}) + \widehat{\Theta}_{i,k}(\mathbf{s}_{t+H}|\mathbf{s}_{t+H}^\star)
    \leq 0, 
    \ \forall i,k, \label{ccp}  
\end{align}
where 
$\widehat{\Theta}_{i,k}(\mathbf{s}_{t+H}|\mathbf{s}_{t+H}^\star)=
\nabla\Theta_{i,k}(\mathbf{s}_{t+H}^\star)(\mathbf{s}_{t+H}-\mathbf{s}_{t+H}^\star)
+\Theta_{i,k}(\mathbf{s}_{t+H}^\star)
$.
By applying the first-order condition for concave functions, we have
$\widehat{\Theta}_{i,k}(\mathbf{s}_{t+H}|\mathbf{s}_{t+H}^\star)
\geq \Theta_{i,k}(\mathbf{s}_{t+H})$.
With the above observation, an inner set of \eqref{xi} can be
obtained if we replace \eqref{xi} by \eqref{ccp} expanded around a feasible point $\mathbf{s}_{t+H}^\star$, which yields
\begin{subequations} 
\begin{align}
&   \mathsf{P}_2:~\min \limits_{\{\mathbf{s}_h, \mathbf{u}_h\}\in\mathcal{F}, \{\mathbf{w},\bm{\xi}\}}~~ 
    \sum_{i=1}^M
  Q_i\|w_i\|_0
    +
   \rho \, C_{\text{tar}}(\{\mathbf{s}_h\})\\
    &~~~~~~~~~~~~~~\text { s.t. }~~\textsf{constraints}~(\ref{collision}), (\ref{w}), (\ref{ccp}).
    \end{align}
\end{subequations} 
Furthermore, since $\widehat{\Theta}_{i,k}(\mathbf{s}_{t+H}^\star|\mathbf{s}_{t+H}^\star)
=\Theta_{i,k}(\mathbf{s}_{t+H}^\star)$ and $\nabla
\widehat{\Theta}_{i,k}(\mathbf{s}_{t+H}^\star|\mathbf{s}_{t+H}^\star)
=\nabla\Theta_{i,k}(\mathbf{s}_{t+H}^\star)$, the gap between $\mathsf{P}_2$ and $\mathsf{P}_1$ goes to zero as $\mathbf{s}_{t+H}\rightarrow\mathbf{s}_{t+H}^\star$ \cite{lipp2016variations,sun2016majorization}.
Denoting the optimal solution to $\mathsf{P}_2$ as $\{\mathbf{s}_h^*,\mathbf{u}_h^*\}$, we 
set $\{\mathbf{s}_h^\star=\mathbf{s}_h^*,\mathbf{u}_h^\star=\mathbf{u}_h^*\}$ as a new expansion point and construct a surrogate set in the next receding-horizon. 
According to the property of CCP, the solution $\{\mathbf{s}_h^*,\mathbf{u}_h^*\}$ to $\mathsf{P}_2$ is local optimal to $\mathsf{P}_1$ \cite{lipp2016variations,sun2016majorization}.

\subsection{Duality-Based Method}

Finally, to tackle the bilevel constraint \eqref{collision} in $\mathsf{P}_2$, we adopt the polyhedron representation and duality-based method in \cite{zhang2020optimization}. 
Specifically, we denote $\mathbb{G}_{h}(\mathbf{s}_h)= \{ \mathbf{x}\in {\mathbb{R}^{3}}| {\mathbf{G}_{h}}(\mathbf{s}_h)\mathbf{x}\preceq 
     {\mathbf{g}_{h}}(\mathbf{s}_t)\}$, and 
$\mathbb{O}_{k}= \{ \mathbf{y}\in {\mathbb{R}^{3}}| {\mathbf{H}_{k}}\mathbf{y}\preceq 
     {\mathbf{h}_{k}}\}
$ as the sets of points for the ego robot and the $k$-th obstacle, respectively, with 
$(\mathbf{G}_{t},\mathbf{g}_{t})$ and 
$(\mathbf{H}_{k},\mathbf{h}_{k})$ defining the rotation and translation of contour lines \cite{li2024edge}. 
Then the inner-level problem of \eqref{collision} is equivalently transformed into its dual form \cite{zhang2020optimization}:
\begin{subequations}
\begin{align}
 \mathsf{D}:\mathop{\mathrm{max}}_{\{\bm{\lambda}_{k,h}\succeq \mathbf{0},\bm{\mu}_{k,h}\succeq \mathbf{0}\}} &
 -{\bm{\lambda}_{k,h}^T}{\mathbf{h}_{k}}- {\bm{\mu}_{k,h}^T}{\mathbf{g}_h}(\mathbf{s}_h) \\
\mathrm{s.t.}~~~~~~ &
    {\left\| {{{\mathbf{H}_{k}}^T}\bm{\lambda}_{k,h} } \right\|} \leq 1,\\
    &{\bm{\mu}_{k,h}^T}{\mathbf{G}_h}(\mathbf{s}_h)
    +{\bm{\lambda}_{k,h} ^T}{\mathbf{H}_{k}}
    =0,
\label{dual}  
\end{align}
\end{subequations}
where $\{\bm{\lambda}_{k,h}$, $\bm{\mu}_{k,h}\}$ are dual variables. 
By solving $\mathsf{D}$ with $\{\mathbf{s}_h\}$ fixed to the solution from the last receding-horizon, the optimal solution $\{\widehat{\bm{\lambda}}_{k,h},\widehat{\bm{\mu}}_{k,h}\}$ is obtained, and \eqref{collision} becomes 
\begin{subequations}
\begin{align}
&-{\widehat{\bm{\lambda}}_{k,h}^T}{\mathbf{h}_{k}}- {\widehat{\bm{\mu}}_{k,h}^T}{\mathbf{g}_h(\mathbf{s}_h)}\geq d_{0},~\forall k,h, \label{dual1} 
\\
&{\widehat{\bm{\mu}}_{k,h}^T}{\mathbf{G}_h(\mathbf{s}_h)}
    +{\widehat{\bm{\lambda}}_{k,h} ^T}{\mathbf{H}_{k}}
    =0,~\forall k,h, \label{dual2}
\end{align}
\end{subequations}
which are convex in ${\mathbf{s}_{h}}$. 
Plugging \eqref{dual1} and \eqref{dual2} into $\mathsf{P}_2$, 
and by approximating $\|w_i\|_0$ as $\|w_i\|_1$ \cite{chamon2019sparse}, problem $\mathsf{P}_2$ becomes a convex optimization problem. 
However, directly conducting joint optimization over $\{\mathbf{s}_{h}, \mathbf{u}_{h},\mathbf{w},\bm{\xi}\}$ may require excessive computation time. 
Thus, we further adopt penalty alternating optimization that iterates between solving $\{\mathbf{s}_{h}, \mathbf{u}_{h}\}$'s subproblem and $\{\mathbf{w},\bm{\xi}\}$'s subproblem. 
Each subproblem is solved by off-the-shelf software such as CVXPY \cite{diamond2016cvxpy}.

\section{Experiments}\label{section4}
\begin{figure*}[t]
\centering
    \begin{subfigure}[t]{0.182\textwidth}
      \includegraphics[width=\textwidth]{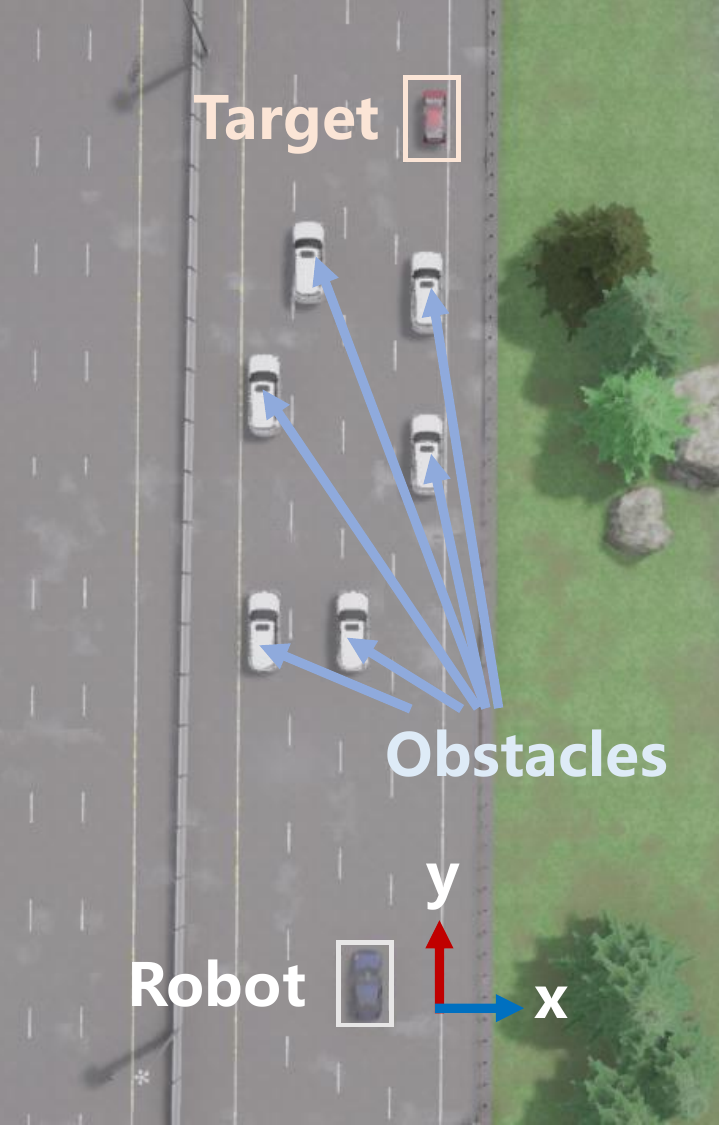}
      \captionsetup{font={small,rm}, labelfont={small,rm}}
      \caption{Setting.}
    \end{subfigure}
    \begin{subfigure}[t]{0.19\textwidth}
        \includegraphics[width=1.00\textwidth]{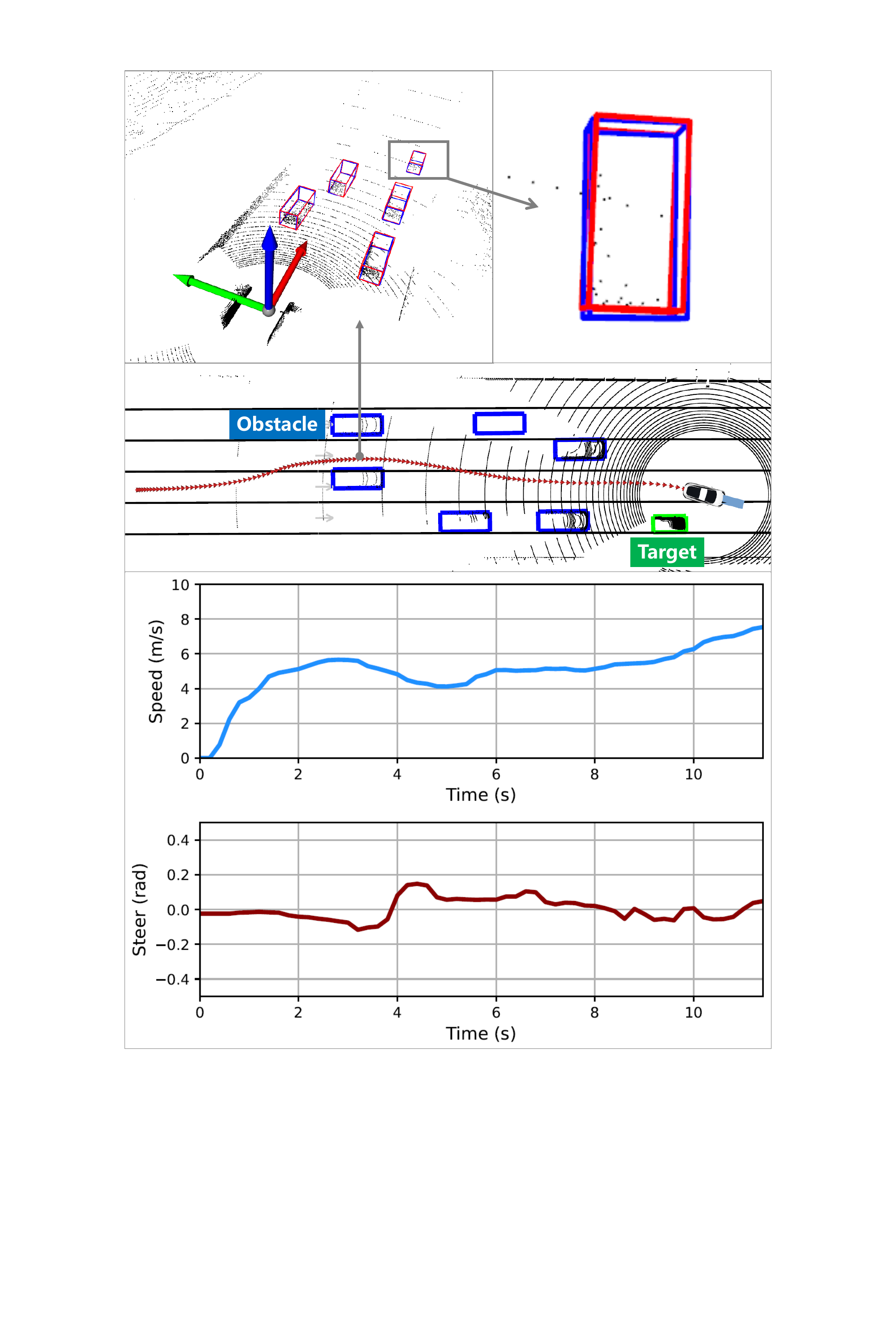}
        \captionsetup{font={small,rm}, labelfont={small,rm}}
        \caption{CROA (ours).}
    \end{subfigure}
    \begin{subfigure}[t]{0.19\textwidth}
        \includegraphics[width=1.00\textwidth]{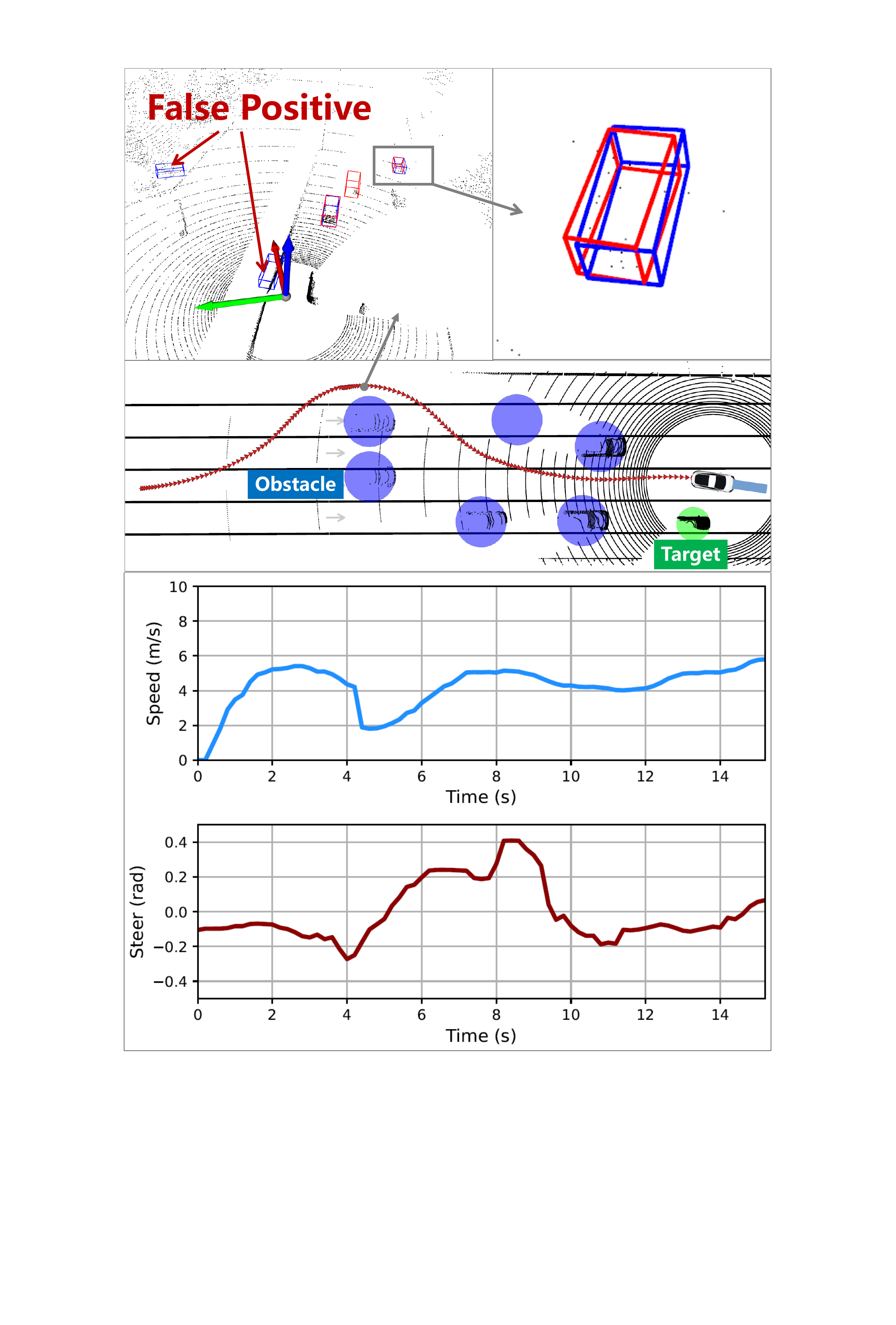}
        \captionsetup{font={small,rm}, labelfont={small,rm}}
        \caption{OMPC.}
    \end{subfigure}
        \begin{subfigure}[t]{0.19\textwidth}
        \includegraphics[width=1.00\textwidth]{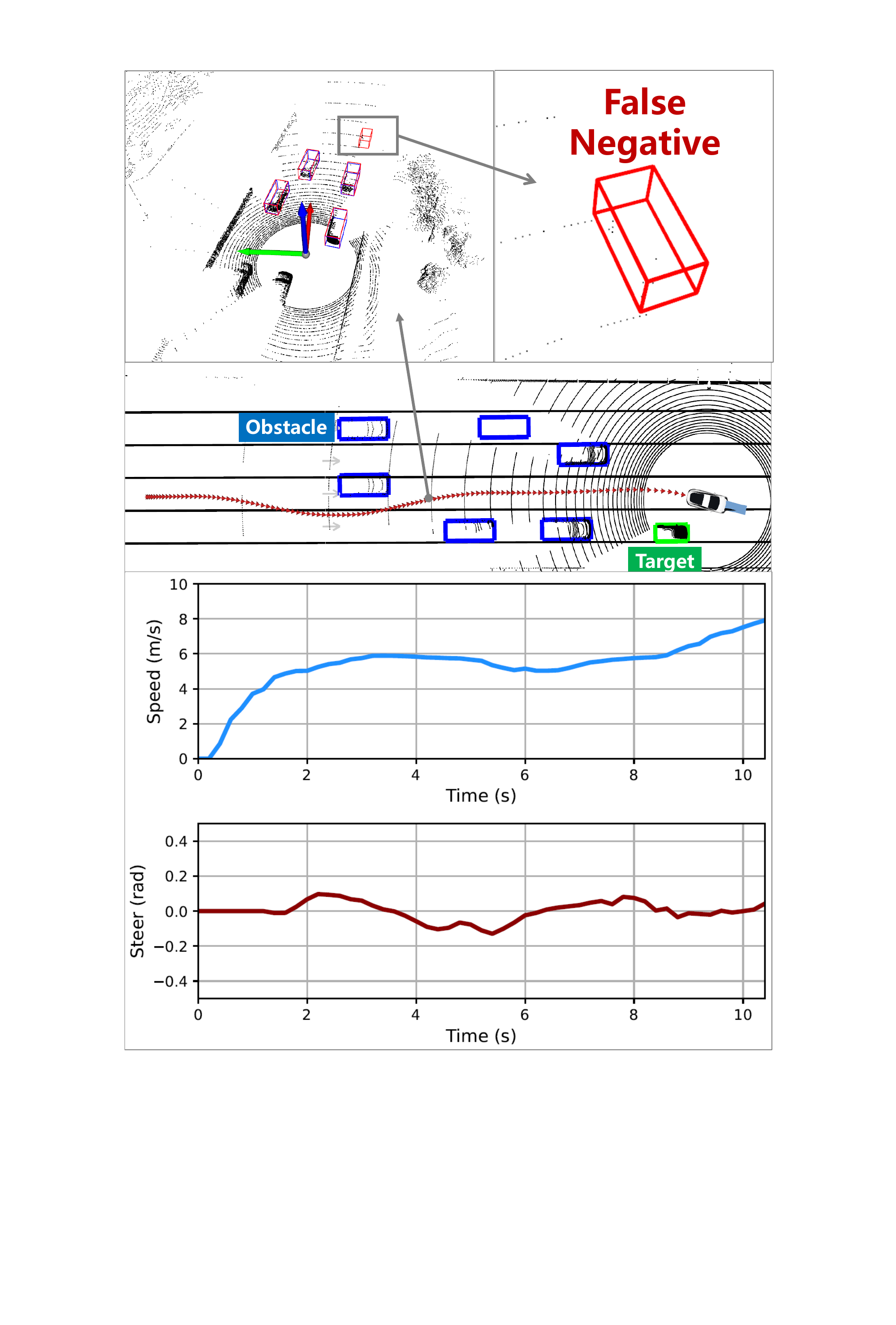}
        \captionsetup{font={small,rm}, labelfont={small,rm}}
        \caption{RDA.}
    \end{subfigure}
        \begin{subfigure}[t]{0.19\textwidth}
        \includegraphics[width=1.00\textwidth]{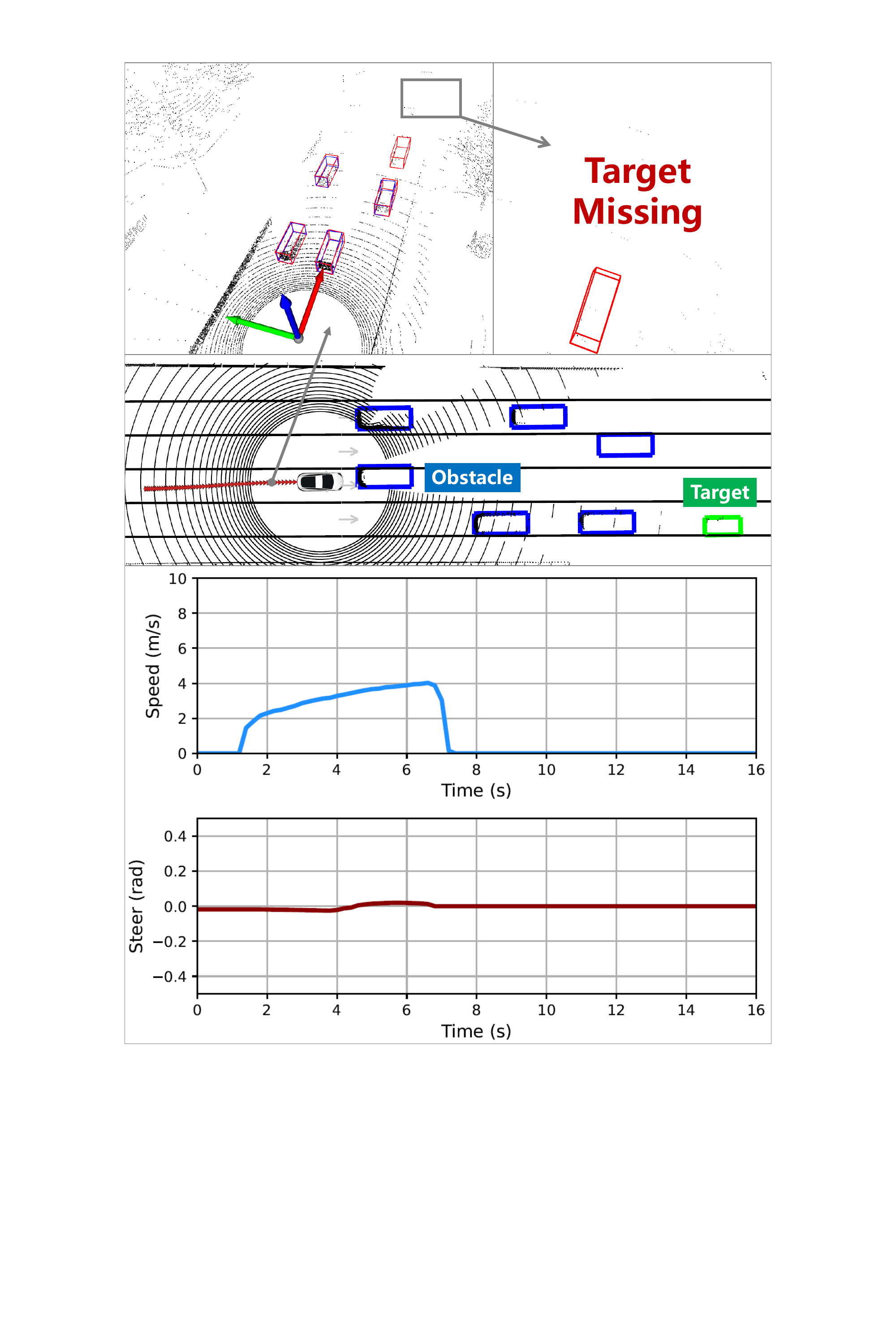}
        \captionsetup{font={small,rm}, labelfont={small,rm}}
        \caption{PF.}
    \end{subfigure}
    \captionsetup{font={small,rm}, labelfont={small,rm}}
    \caption{{Settings and motion profiles. In upper figures of (b)--(e), ground-truth and detected boxes are marked in red and blue, respectively. In middle figures of (b)--(e), robot trajectories, obstacles, and target are marked as red arrow lines, blue boxes, and green boxes, respectively.}
    }
\end{figure*}

\begin{figure}[t]
\centering
    \begin{subfigure}[t]{0.75\textwidth}
      \includegraphics[width=\textwidth]{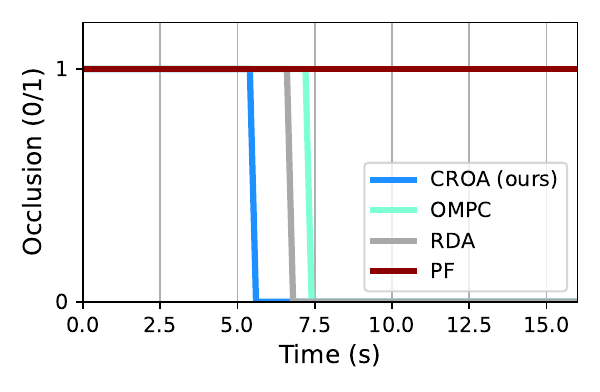}
      \captionsetup{font={small,rm}, labelfont={small,rm}}
      \caption{Occlusion states.}
    \end{subfigure}
    ~
    \begin{subfigure}[t]{0.75\textwidth}
  \includegraphics[width=\textwidth]{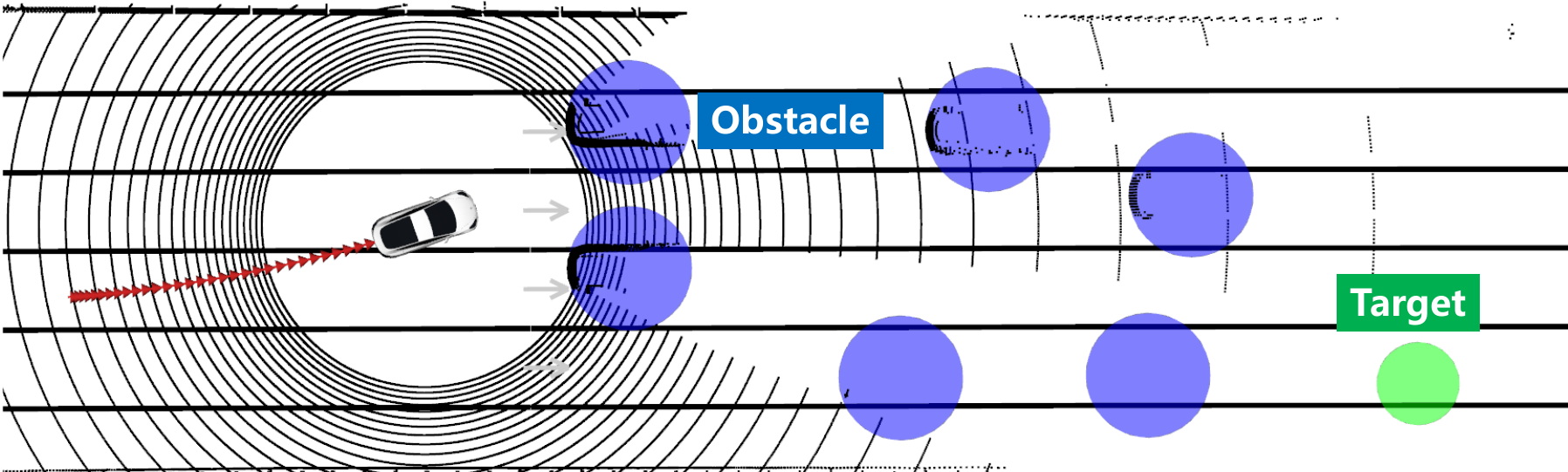}
  \captionsetup{font={small,rm}, labelfont={small,rm}}
  \caption{OMPC gets stuck occasionally.}
\end{subfigure}
    \captionsetup{font={small,rm}, labelfont={small,rm}}
    \caption{{Occlusion states of Fig. 2b--2e and OMPC failure.}
    }
\end{figure}

\begin{figure}[t]
\centering
    \begin{subfigure}[t]{0.75\textwidth}
      \includegraphics[width=\textwidth]{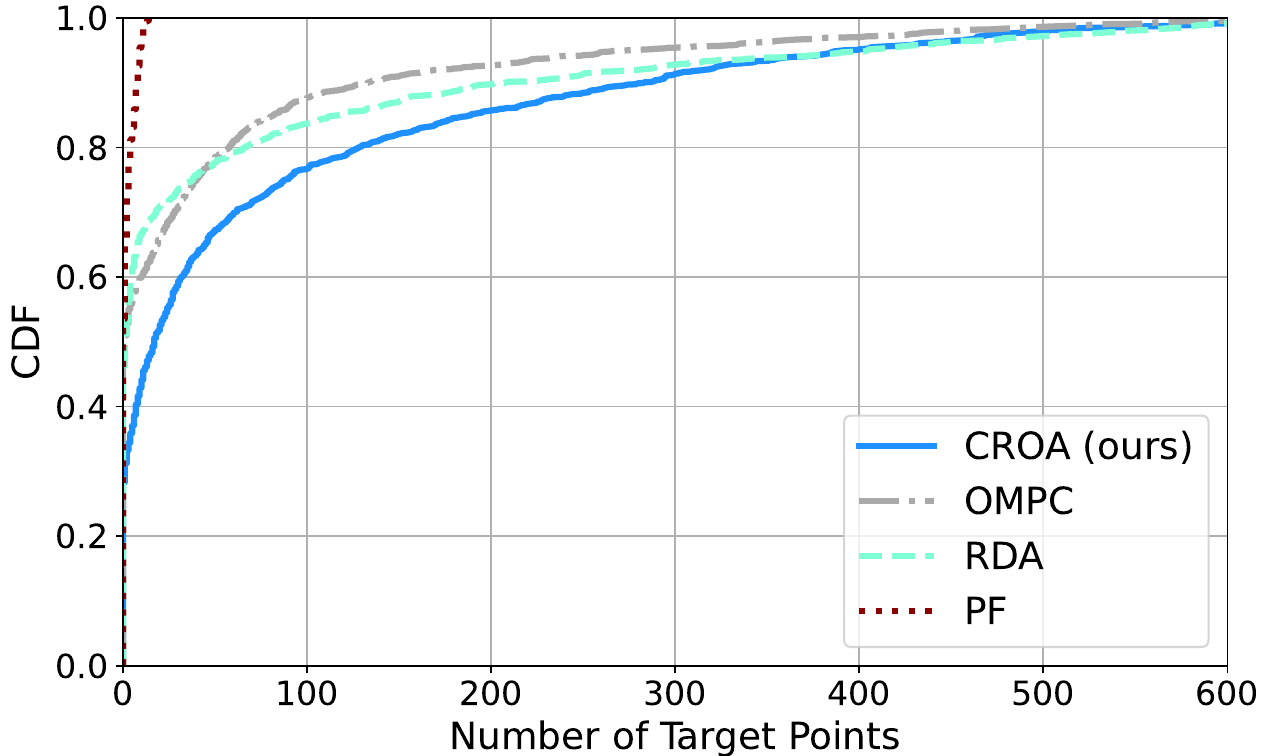}
      \captionsetup{font={small,rm}, labelfont={small,rm}}
      \caption{CDF of target points.}
    \end{subfigure}
    ~
    \begin{subfigure}[t]{0.75\textwidth}
  \includegraphics[width=\textwidth]{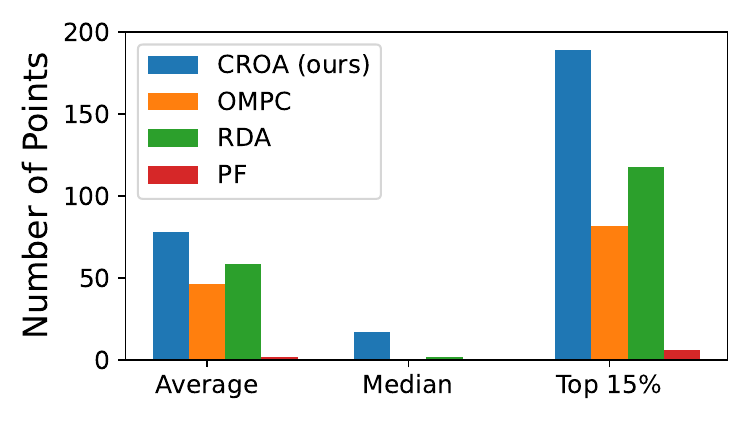}
  \captionsetup{font={small,rm}, labelfont={small,rm}}
  \caption{Number of target points.}
\end{subfigure}
    \captionsetup{font={small,rm}, labelfont={small,rm}}
    \caption{{Comparison of view qualities.}
    }
\end{figure}

We implement CROA using Python in ROS.
Evaluations are conducted in a multi-lane urban driving scenario in Town04 of Carla \cite{dosovitskiy2017carla} with $K=6$ as shown in Fig. 3a. 
The safety distance is set to $d_0=1$\,m.
The length of prediction horizon is set to $H=10$, with a time step of $\Delta t = 0.3$\,s.
The robot is a Tesla Model3 equipped with a $64$-line lidar at $10\,$Hz with vertical field of view being $\pm13.45^{\circ}$. 
The DNN detector is sparsely embedded convolutional detection (SECOND) network \cite{yan2018second}, trained by $5000$ point-cloud frames in Carla Town02 for $50$ epochs. 
The longitudinal and lateral wheelbases for computing vehicle dynamics $f(\cdot)$ are $2.87$\,m and $1.75$\,m, respectively. 
The target is an Audi A2, and obstacles are Nissan Patrols. 
Vehicle positions are marked in Fig. 3a, with $\pm2\,$m random variations along $x$ and $y$ axes. Each point in the figure is obtained by averaging over $20$ simulation runs, with independent realizations in each run. The target is defined as occluded (not detectable) if the number of lidar points on its body is less than $10$, and non-occluded (detectable) otherwise \cite{zhang2021distributed, geiger2013vision}.

We compare CROA to the following schemes: 1) PF-assisted detection (PF for short) \cite{hung2023review,yan2018second}, which is a baseline combining PF \cite{hung2023review} and SECOND \cite{yan2018second}; 2) RDA-assisted detection (RDA for short) \cite{han2023rda}, which is a state-of-the-art PMAD scheme; 3) OMPC-assisted detection (OMPC for short) \cite{nageli2017real}, which is an occlusion-aware DMAD scheme that ignores obstacle geometries.

Detection, trajectory, and control profiles of all schemes are shown in Fig. 3b--3e. 
As seen from Fig. 3b, the proposed CROA successfully finds the narrow occlusion-free path between two close obstacles. 
At $t=4\,$s, the robot steers to the left and soon sees the target at $t=5.5\,$s as shown in Fig. 4a. 
The detection becomes accurate at $t=6\,$s as shown in the upper part of Fig. 3b.
The average speed is about $5\,$m/s, allowing the robot to reach the target within $11\,$s.

For OMPC, the robot shifts to the leftmost lane, since the occlusion cost at the leftmost lane is smallest under the distance-based occlusion model. 
As seen from its control profile, the robot adopts a sharp left turn from $t=5\,$s to $t=9\,$s. 
This results in detour and longer time to reach the target (i.e., $15\,$s).
Views along the path also become worse: the robot sees the target at $t=7.5\,$s ($2$ seconds later than CROA) as shown in Fig. 4a, and the detected box at $t=8\,$s does not accurately match the ground truth as shown in upper part of Fig, 3c (there also exist false positives).

For RDA, it finds the shortest path to reach the target (i.e., $10.3\,$s). 
This is realized by steering to the right at $t=4\,$s, as shown in its control profile, which is the opposite direction of CROA. 
However, the lidar data from $t=4\,$s to $t=7\,$s is severely occluded. 
For example, the detector at $t=7.5\,$s misses the target due to insufficient amount of lidar points. 

Lastly, PF brakes behind the obstacle, observing no target points at all in the illustrated case. 
Note that under random variations of scenarios, the PF scheme may partially observe the target. The OMPC scheme may get stuck in certain realizations as shown in Fig. 4b, due to its conservative collision model that represents each obstacle as a ball. 
This demonstrates the necessity of adopting accurate obstacle models for achieving clutter-resilience.

\begin{table}[t]
\centering
\caption{Comparison of detection performance.}
\renewcommand{\arraystretch}{1.32} 
\begin{center}
\begin{tabular}{c|cccc}
\hline
\multicolumn{1}{c|}{\multirow{2}{*}{Method}}                                                       & \multicolumn{1}{c}{Detectable} & \multicolumn{1}{c}{Occlusion} 
& \multicolumn{2}{c}{AP\_R40$\text{@}$IoU=0.7 $\uparrow$ }
\\ \cline{4-5} 
\multicolumn{1}{c|}{}   & \multicolumn{1}{c}{Frames $\uparrow$ } &  \multicolumn{1}{c}{Ratio $\downarrow$} 
 & Moderate    & Hard   \\ \hline
 CROA & \textbf{764} & \textbf{46.7\%} & \textbf{82.5\%}   & \textbf{75.0\%}  \\ 
 \cline{1-5} 
 OMPC     & $542$ ($\downarrow 222$) & $61.3\%$  ($\uparrow 14.6\%$) & $77.5\%$ ($\downarrow 5.0\%$)   &75.0\% ($0.0\%$) \\ 
 \cline{1-5} 
RDA & $352$ ($\downarrow 412$) & $68.0\%$ ($\uparrow 21.3\%$) & 80.0\% ($\downarrow 2.5\%$)   & 70.0\% ($\downarrow 5.0\%$) \\ 
\cline{1-5} 
PF & $45$ ($\downarrow 719$) & $97.1\%$  ($\uparrow 50.4\%$)  & 0.0\% ($\downarrow 82.5\%$)   & 0.0\% ($\downarrow 75.0\%$) \\ \hline
\end{tabular}
\end{center}
\label{table.rsfl_vs_basp}
\end{table}

The cumulative distribution function (CDF) of the number of observed points on the target is shown in Fig. 5a. The robot collects the most measurements of the target under CROA, since its CDF curve is consistently below the other CDF curves as the number of points increases from $0$ to $600$.
The average, median, top $15\%$ number of points of CROA in Fig. 5b are at least $34.3\%$, $700.0\%$, $60.3\%$ larger than all the other methods. This demonstrates that CROA actively plans an executable path with reduced occlusion. 

Detection performances of different schemes are shown in Table 1. 
The proposed CROA achieves the smallest occlusion ratio, with at least $14.6\%$ improvement compared to all the other schemes. Consequently, the number of detectable frames under CROA is the largest (i.e., $764$).
Furthermore, for detectable frames, the average precision (AP) score at IoU=0.7 (IoU is intersection over union) \cite{geiger2013vision,zhang2021distributed} of CROA is also the highest for both moderate and hard instances. 
This implies that the total number of valid detections for the target under CROA is significantly larger than that of other benchmarks.

\section{Conclusion}\label{section5}
This paper presented CROA for active object detection in clutter.
Viewpoints of the ego robot were optimized to minimize the occlusion probability under vehicle dynamics. Based on accurate occlusion and collision models, and by leveraging nonconvex optimization techniques, CROA established a higher capability of finding paths with good views and no detour. The CROA was implemented in ROS and evaluated in Carla. 
Experimental results demonstrated the effectiveness and robustness of CROA. 


\bibliographystyle{IEEEtran}

\end{document}